\documentclass[11pt,a4paper]{article}
\usepackage[hyperref]{acl2021}
\usepackage{times}
\usepackage{latexsym}

\usepackage{subcaption}
\usepackage{graphicx}
\usepackage{booktabs}
\usepackage{eqparbox}
\usepackage{arydshln}
\usepackage{multirow}
\usepackage{amsmath}
\usepackage{enumitem}
\usepackage{array}
\usepackage{float}
\usepackage{soul}
\usepackage{microtype}

\usepackage{microtype}

\aclfinalcopy 

\title{Constructing Flow Graphs from Procedural Cybersecurity Texts}

\newcommand*\samethanks[1][\value{footnote}]{\footnotemark[#1]}

\author{
Kuntal Kumar Pal\thanks{\quad These authors contributed equally to this work.} ,
Kazuaki Kashihara\samethanks,
Pratyay Banerjee\samethanks,\\
\textbf{Swaroop Mishra,
Ruoyu Wang,
Chitta Baral}\\
{School of Computing, Informatics, and Decision Systems Engineering,}\\
{Arizona State University},\\
\{kkpal, kkashiha, pbanerj6, srmishr1, fishw, chitta\}@asu.edu}

\date{}

\begin{document}
\maketitle
\begin{abstract}
Following procedural texts written in natural languages is challenging.
We must read the whole text to identify the relevant information or identify the instruction-flow to complete a task, which is prone to failures.
If such texts are structured, we can readily visualize instruction-flows, reason or infer a particular step, or even build automated systems to help novice agents achieve a goal.
However, this structure recovery task is a challenge because of such texts' diverse nature.
This paper proposes to identify relevant information from such texts and generate information flows between sentences.
We built a large annotated procedural text dataset (CTFW) in the cybersecurity domain (3154 documents).
This dataset contains valuable instructions regarding software vulnerability analysis experiences.
We performed extensive experiments on CTFW with our LM-GNN model variants in multiple settings.
To show the generalizability of both this task and our method, we also experimented with procedural texts from two other domains (Maintenance Manual and Cooking), which are substantially different from cybersecurity.
Our experiments show that Graph Convolution Network with BERT sentence embeddings outperforms BERT in all three domains.

\end{abstract}

\section{Introduction}


Many texts in the real-world contain valuable instructions.
These instructions define individual steps of a process and help users achieve a goal (and corresponding sub-goals).
Documents including such instructions are called \emph{procedural texts}, ranging from simple cooking recipes to complex instruction manuals.
Additionally, \textit{discussion in a shared forum or social media platform, teaching books, medical notes, sets of advice about social behavior, directions for use, do-it-yourself notices, itinerary guides can all be considered as procedural texts} \cite{delpech-saint-dizier-2008-investigating}. 
Most of these texts are in the form of natural languages and thus, lacking structures.
We define \emph{structure} as sentence-level dependencies that lead to a goal.
These dependencies can vary based on the text-domain.
Some examples of such dependencies are action traces, effects of an action, information leading to the action, and instruction order.
Constructing structured flow graphs out of procedural texts is the foundation for natural language understanding and summarization, question-answering (QA) beyond factoid QA, automated workflow visualization, and the recovery of causal relationships between two statements. By flow-graph we mean both information and action flows in a text.
However, the lack of structures in such texts makes them challenging to follow, visualize, extract inferences, or track states of an object or a sub-task, which ultimately makes constructing their flow graphs an insurmountable task.


Procedural texts are common in cybersecurity, where security analysts document how to discover, exploit, and mitigate security vulnerabilities in articles, blog posts, and technical reports, which are usually referred to as \emph{security write-ups}.
Practitioners in cybersecurity often use write-ups as educational and researching materials.
Constructing structured flow graphs from security write-ups may help with automated vulnerability discovery and mitigation, exploit generation, and security education in general.
However, automatically analyzing and extracting information from security write-ups are extremely difficult since they lack structure.

\begin{figure}[t]
  \includegraphics[width=.95\linewidth]{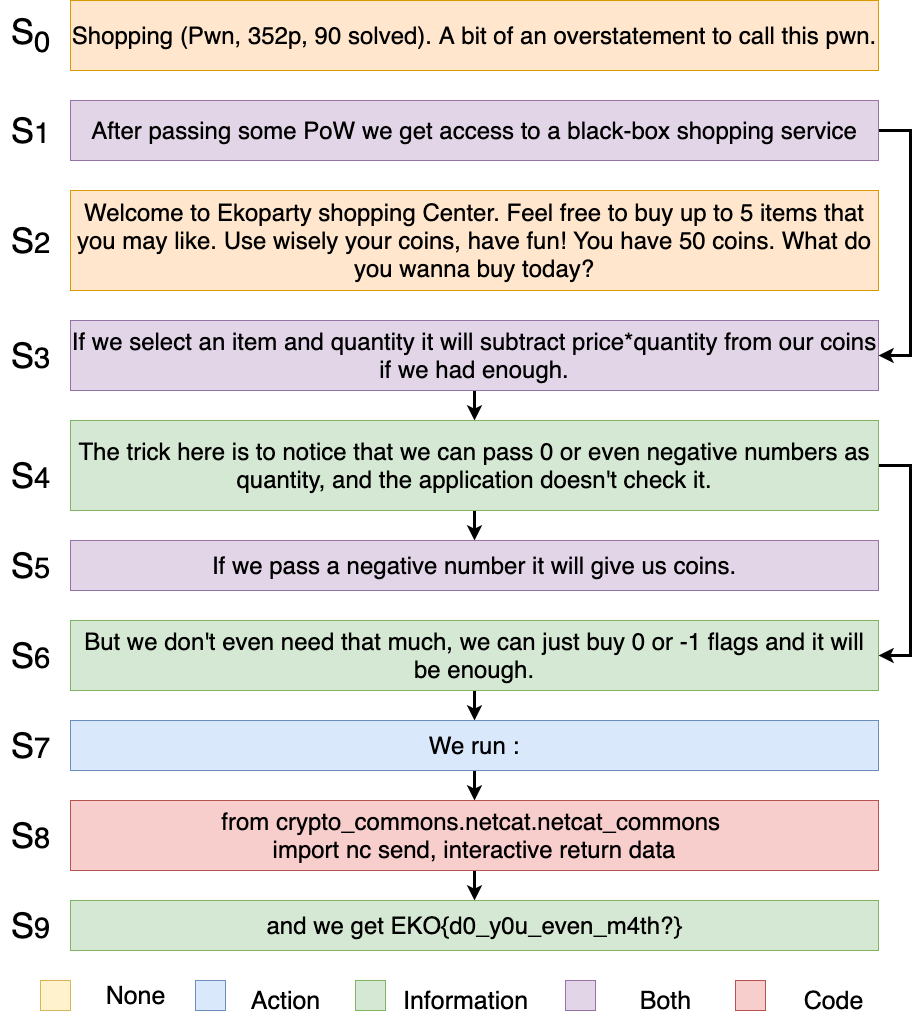}
  \caption{An example flow graph from the CTFW. Sentences in $S_2$ are merged into one block for clarity.}
  \label{fig:ctf_example}
\end{figure}

Figure \ref{fig:ctf_example} illustrates the core of a security write-up (broken into sentences) that carries instructions for exploiting a vulnerability in an online shopping service.
$S_1$, $S_3$, and $S_4$ are the author's observations about the service's nature.
Based on this information, $S_5$ and $S_6$ are two possible paths of actions.
The author chose $S_6$ and ran the Python code in $S_8$ to exploit the service.
$S_0$ and $S_2$ are irrelevant for the author's goal of exploiting this service.

Here we propose a novel approach to extract action paths out of structure-less, natural language texts by identifying actions and information flows embedded in and between sentences and constructing action flow graphs.
Specifically, our focus is on procedural texts in the cybersecurity domain.
We also show that constructing flow graphs helps extract paths of actions in domains besides cybersecurity, such as cooking and maintenance manuals.

Most previous works \cite{ mori2014flow, kiddon2015mise, malmaud2014cooking, maeta2015framework, xu2020benchmark, mysore2019materials,song2011procedural} focus on fine-grained knowledge extraction from procedural texts in diverse domains.
There are also a handful of works \cite{delpech-saint-dizier-2008-investigating,fontan2008analyzing,jermsurawong2015predicting} that study the structure of natural language texts.
Different from previous works, we extract structures and construct flow graphs from natural texts at the sentence level.
This is because fine-grained domain-entity extraction tasks require a large amount of annotated data from people with specific in-depth domain knowledge, whereas text structures can be generalized.

\noindent
\textbf{Dataset.}
We built a dataset from security write-ups that are generated from past Capture The Flag competitions (CTFs).
CTFs are computer security competitions that are usually open to everyone in the world.
Players are expected to find and exploit security vulnerabilities in a given set of software services, and through exploiting vulnerabilities, obtain a \emph{flag}---a unique string indicating a successful  attempt---for each exploited service.
Once the game is over, many players publish security write-ups that detail how they exploited services during the game.
While these write-ups are a valuable educational resource for students and security professionals, they are usually unstructured and lacking in clarity.
We collected 3617 CTF write-ups from the Internet, created a procedural text dataset, and invited domain experts to label each sentence for the purpose of constructing flow graphs and identifying action paths.
To the best of our knowledge, this is the first attempt to use the knowledge embedded in security write-ups for automated analysis. The data and the code is publicly available \footnote{https://github.com/kuntalkumarpal/FlowGraph} for future research.




\medskip
\noindent
This paper makes the following contributions:
\begin{itemize}[noitemsep,nosep,leftmargin=1.5em]
    \item
    We built a new procedural text dataset, CTFW, in the cybersecurity domain.
    To the best of our knowledge, CTFW is the first dataset that contains valuable information regarding vulnerability analysis from CTF write-ups.

    \item We proposed a new NLU task of generating flow graphs from natural language procedural texts at the sentence level without identifying fine-grained named entities.

    \item We proposed four variations of a graph neural network-based model (LM-GNN) to learn neighbor-aware representation of each sentence in a procedural text and predict the presence of edges between any pair of sentences.
    
    \item We evaluated our models on CTFW.
    To the best of our knowledge, this is the first attempt in automated extraction of information from security write-ups.
    We also evaluated our models across three datasets in different domains and showed the generalizability of our approach. 
\end{itemize}

\section{Our Approach}

We map each sentence of a procedural text as a node in a graph, and the action or information flows as edges. The task is then simplified into an edge prediction task:
Given a pair of nodes, find if there is an edge between them.
We learn feature representations of nodes using language models like BERT/RoBERTa \cite{devlin2018bert,liu2019roberta}.
Then, to make the nodes aware of their neighboring sentences, we use Graph Neural Network (GNN) to update the node representations.
We check for the edge between every pair of nodes in a graph and reduce the task to a binary classification during inference. This  formulation enables us to predict any kind of structure from a document.



\section{Dataset Creation}

\begin{table}[]
\begin{tabular}{@{}lccc@{}}
\toprule
\textbf{Dataset Statistics}             & \textbf{COR} & \textbf{MAM} & \textbf{CTFW} \\ \midrule
\# Documents             &  297   &    575   &  3154    \\ 
Avg size of document     &  9.52   &    8.12&    17.11  \\ 
Avg length of sentence   &   65.46  &   34.81&     92.87 \\ 
\# Edges ($|e^+|$)              &   2670  &    5043&     54539 \\ 
 $|e^+|:(|e^+| + |e^-|)$    &  0.18   &    0.12&  0.07    \\ 
Avg degree of node       &  1.83   &    1.76&    1.88  \\ \bottomrule
\end{tabular}
\caption{Dataset Statistics. $|e^+|$ is the total number of actual edges, and $|e^+| + |e^-|$ is the total number of edges possible. The in-degree of the starting node and out-degree of the end node are both 0.}
\label{tab:stats}
\end{table}

In this section, we present how we created three datasets on which we evaluated our approach. Table \ref{tab:stats} shows the statistics for each datasets used.

\subsection{CTF Write-ups Dataset  (CTFW)}
\label{sub:CTFW}
Each CTF competition has multiple challenges or tasks. Each task may have multiple write-ups by different authors. 
We crawled 3617 such write-ups from GitHub and CTFTime~\cite{ctftime}.
Write-ups are unique and diverse but have common inherent principles.
For each write-up, we provide two kinds of annotations: \textit{sentence type} and \textit{flow structure}.
The writing style is informal with embedded code snippets and often contains irrelevant information.

Part of the annotations were provided as an optional, extra-credit assignment for the Information Assurance course. These CTF write-ups were directly related to the course-content, where students were required to read existing CTF write-ups and write write-ups for other security challenges they worked on during the course. Then students were given the option of voluntarily annotating CTF write-ups they read for extra credits in the course. For this task, we followed all the existing annotation guidelines and practices. We also ensured that (1) The volunteers were aware of the fact that their annotations would be used for a research project (2) They were aware that no PII was involved or would be used in the research project (3) They were aware that extra credits were entirely optional, and they could refrain from submitting at any point of time without any consequences (4) Each volunteer was assigned only 10-15 write-ups based on a pilot study we did ahead of time, annotating an average-length CTF write-up took about two minutes (maximum ten mins).

Remaining annotations were performed by the Teaching Assistants (TA) of the course. These annotations were done as part of the course preparation process, which was part of their work contract. All the TAs were paid bi-weekly compensation by the university or by research funding. It was also ensured that the TAs knew these annotations would be used for a research project, their PII was not involved and annotations were to be anonymized before using. We verified the annotations by randomly selecting write-ups from the set. Figure \ref{fig:ctf_example} shows a sample annotation.

\noindent\textbf{Sentence Type Annotations.}
We split the documents into sentences using natural language rules. We then ask the volunteers to annotate the type of each statement as either Action (A), Information (I), Both (A/I), Codes (C), or irrelevant (None). \textit{Action} sentences are those where the author specifies actions taken by them, whereas, \textit{Information} statements mention observations of the author, the reasons and effects of their action. Sentences containing \textit{codes} are assigned as C, and those which can be considered as both information and actions are marked as \textit{Both} (A/I).

\noindent\textbf{Flow structure Annotations.}
The second level of annotations is regarding the write-up structure. Each volunteer is given a csv file for each document with a set of sentence IDs and text for each write-up. They are asked to annotate the flow of information in the document by annotating the sentence id of some next possible sentences, which indicate the flow. 
We filter those write-ups which are irrelevant and those which did not have much detail (single line of valuable information). We call a write-up as irrelevant if it has no action-information annotations or if it has direct codes without any natural language description of steps to detect vulnerabilities.
We only keep write-ups written in the English language for this work. Finally, we have 3154 write-ups with \textit{sentence type}  and \textit{structure annotations}.

CTFTime website states that the write-ups are copyrighted by the authors who posted them and it is practically impossible to contact each author. Such data is also allowed to use  for academic research purposes\cite{uscopyright,eucopyright}.
Thus, we follow the previous work using data from CTFTime \cite{vsvabensky2021cybersecurity}, and share only the urls of those write-ups which we use. We do not provide the scraper script since it would create a local copy of the write-up files unauthorized by the users. Interested readers can replicate the simple scraper script from the instructions in Appendix \ref{sub:appendix:dataset} and use it after reviewing the conditions under which it is permissible to use. We, however, share our annotations for those write-up files.

\subsection{Cooking Recipe Flow Corpus (COR)}
This corpus \cite{yamakata2020english} provides 300 recipes with annotated recipe named entities and fine-grained interactions between each entity and their sequencing steps. Since we attempt to generate action flow graphs without explicitly identifying each named entity, we aggregate the fine-grained interactions between recipe named entities to generate sentence-level flows for each recipe. We reject three single-sentence recipes.

\subsection{Maintenance Manuals Dataset  (MAM)}
This dataset \cite{qian2020approach} provides multi-grained process model extraction corpora for the task of extracting process models from texts. It has over 160 Maintenance Manuals. Each manual has fine-grained interactions between each entity and its sequencing steps. We use the annotations from sentence-level classification data and semantic recognition data for generating annotations of sentence-level flows for each process. 
Here also, we reject single sentence processes. 


\section{Model Description}
 Our goal is to find paths or traces of actions or information between texts. This needs an understanding of each sentence's interconnection. Hence, we modeled the problem into an edge prediction task in a graph using GNNs.  We represent each sentence as a node and directed edges as information flows. Since this is procedural text (unidirectional nature) of instructions, we consider only the directed edges from one sentence $S_n$ to any of its next sentences $S_{n+i}$. The node representations are learned using language models (LM) and GNNs.

\subsection{Document to Sentence Pre-processing}
Given a  natural language document, first we split the document into sentences based on simple rules and heuristics. COR and MAM datasets already have document split into separate sentences. In the flow graph creation task, we filter out irrelevant sentences for the CTFW dataset based on the sentence type annotations. After this pre-processing task, each document ($D_i$) is converted into a series of sentences ($S_j$) where $n$ is the number of valid sentences in a document.

\begin{equation*}
    D_i = \lbrace S_0, S_1, S_2 ... S_{n-1} \rbrace
\end{equation*}

\begin{figure}[ht]
  \includegraphics[width=.95\linewidth]{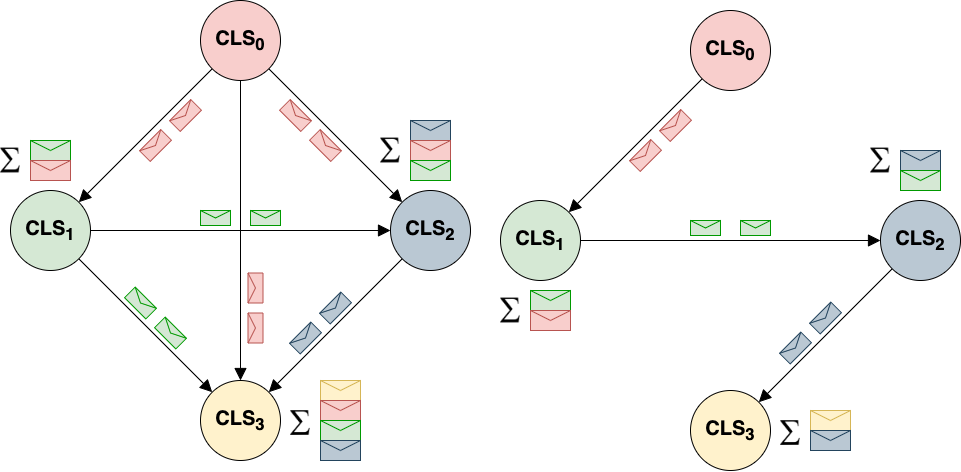}
  \caption{\textit{Node Representation Learning for a document with four sentences in single-layer GNN}. Left: \textit{Semi-Complete} Structure, Right: \textit{Linear} Structure. During training, the sentence representation ($\mathit{CLS}_i$) are enriched using appropriate message passing techniques from the connected 1-hop neighbors.}
  \label{fig:model}
\end{figure}
\subsection{Document to Graph Representation}
A graph  ($G =  (V,E) $) is formally represented as a set of nodes  ($V=\{v_0, v_1, ..\}$) connected by  edges  ($E = \{e_0, e_1, ..\}$ where $e_i=\{v_m, v_n\}$). We consider the sentences  ($S_j$) of any document  ($D_i$) as nodes of a directed graph  ($G_i$). 
We experiment with two graph structure types for learning better node representation using GNN. 
First, we form local windows ($W_N$, where $N=3,4,5,all$ sentences) for each sentence and allow the model to learn from all of the previous sentences in that window.
 We form the document graph by connecting each sentence with every other sentence in that window, with directed edges only from $S_i$ to $S_j$ where $i < j$. We do this since procedural languages are directional. We call this configuration \textit{Semi-Complete}. Second, we consider connecting the nodes linearly where every $S_i$ is connected to $S_{i+1}$ except the last node. We call this  \textit{Linear} setting. Figure \ref{fig:model} shows the settings. 
We use LMs like BERT and RoBERTa to generate initial sentence representations. For each sentence  ($S_i$), we extract the pooled sentence representation ($\mathit{CLS}_{S_i}$) of contextual BERT/RoBERTa embeddings  ($h_{S_i}$). We use $\mathit{CLS}_{S_i}$ as node features for the graph  ($G_i$).
\begin{equation*}
    h_{S_i} = \mathit{BERT}\,  (\, [\mathit{CLS}]s_0s_1...s_{n-1}\,[\mathit{SEP}]\, )
\end{equation*}

\subsection{Neighbor Aware Node Feature Learning} Since the LM sentence vectors are generated individually for each sentence in the document, they are not aware of other local sentences. So, through the \textit{semi-complete} graph connection, the model can learn a global understanding of the document. However, the \textit{linear} connection helps it learn better node representation conditioned selectively on its predecessor. We call the connected nodes as the neighbor nodes. 
We use Graph Convolutional Network (GCN) \cite{kipf2016semi} and Graph Attention Network  (GAT) \cite{velivckovic2017graph} to aggregate the  neighbor information for each node following the generic graph learning function \eqref{eq1}
\begin{equation}
    \mathbf{H}^{l+1} = f (\mathbf{H}^l, \mathbf{A}) 
    \label{eq1}
\end{equation}
where $\mathbf{A}$ is the adjacency matrix of the graph, $\mathbf{H}^l$ and $\mathbf{H}^{ (l+1)}$ are the node representations at $l$th and $ (l+1)$th layer of the network and $f$ is the message aggregation function. In GCN, each node $i$, aggregates the representations of all of its neighbors $N (i)$ based on $\mathbf{A}$ and itself at layer $l$ and computes the enriched representation $\mathbf{h}_i^{l+1}$ based on the weight matrix $\mathbf{\Theta}$ of the layer normalized by degrees of source $d(i)$ and its connected node $d(j)$ as per \eqref{eq2}. In GAT,  messages are aggregated based on  multi-headed attention weights  ($\alpha$) learned from the neighbor node representations $\mathbf{h}^l_j$ following \eqref{eq3}.
\begin{align}
    \mathbf{h}_i^{l+1}&=\mathbf{\Theta}\sum_{j\in N(i)\cup\{i\}}\frac{1}{\sqrt{d(i)d(j)}}\mathbf{h}_j^{l} \label{eq2}\\
    \mathbf{h}_i^{l+1}&=\alpha_{ii}\mathbf{\Theta}\mathbf{h}^l_i \quad+\sum_{j\in N(i)}\alpha_{ij}\mathbf{\Theta}\mathbf{h}^l_j \label{eq3}
\end{align}

\subsection{Projection}
We concatenate the neighbor aware node representations of each pair of nodes ($\mathbf{h}_i$;$\mathbf{h}_j$) from a graph and pass it through two projection layers with a GELU \cite{hendrycks2016gaussian} non-linearity in between. We use the same non-linearity functions used by the BERT layers for consistency. We steadily decrease the parameters of each projection layer by half. During testing, given a document, we are unaware of which two sentences are connected. So, we compare each pair of nodes. This leads to an unbalanced number of existing (1) and non-existing (0) edge labels. Hence, we use weighted cross-entropy loss function as in equation \eqref{eq4} and \eqref{eq5}, where $L$ is the weighted cross-entropy loss, $w_c$ is the weight for class $c$, $i$ is the data in each mini-batch.
\begin{align}
    L(x,c) &= w_{c}\Big(-x_c + log\Big(\sum_j exp(x_j)\Big)\Big)
    \label{eq4}\\
    L &= \frac{\sum_{i=1}^N L(i,c_{i})}{\sum_{i=1}^N w_{c_i}}
    \label{eq5}
\end{align}

\subsection{Training and Inference}
Our training data comprises a set of sentences and the connections as an adjacency matrix for each document. Batching is done based on the number of graphs. GCN/GAT updates the sentence representations. A pair of node representations are assigned a label of 1 if there is an edge between them; otherwise, we assign them 0.
Thus, we model it as a binary classification task as in equation \eqref{eq6} where $f$ is the projection function, $g$ is the softmax function, and $y$ is the binary class output. Depending on the weighted cross-entropy loss, the node representations get updated after each epoch. During inference, the model generates node representations of each sentence in a test document, and we predict whether an edge exists between any two nodes in a given document graph.
\begin{align}
    y_c = \arg\max_{k} g ( f (\mathbf{h}_i;\mathbf{h}_j),k) \quad c\in\lbrace0,1\rbrace
    \label{eq6}
\end{align}

\section{Experiments}
\textbf{Datasets and Tasks:}
Each dataset is split into train, validation, and test sets in 70:10:20 ratio.
The first task is identifying relevant information from raw CTF write-ups by classifying the type of each sentence.
The second task is identifying information flows between sentences by predicting edges between sentence pairs, if any.

\begin{table*}[]
\centering
\small
\begin{tabular}{@{}llcccccc@{}}
\toprule
\multicolumn{2}{c}{\multirow{2}{*}{Models}} & \multicolumn{2}{c}{CTFW}                            & \multicolumn{2}{c}{COR}                             & \multicolumn{2}{c}{MAM}                             \\ \cmidrule(l){3-8} 
\multicolumn{2}{c}{}                        & \multicolumn{1}{c}{PRAUC} & \multicolumn{1}{c}{F1} & \multicolumn{1}{c}{PRAUC} & \multicolumn{1}{c}{F1} & \multicolumn{1}{c}{PRAUC} & \multicolumn{1}{c}{F1} \\ \midrule
\multirow{4}{*}{Baselines}     & Random            &                                 -          &      50.49                   & -
            &                   42.78&
-           &                  47.82        \\
    & Weighted Random   &      -                      &            37.81             &          -                              &         39.13                &            
-                &                 44.10        \\
    & BERT-NS           &                        0.5751    &          26.12               
                &       \underline{0.5638}       
                &             43.14            
                &   
0.5873        &                    29.73     \\
    & RoBERTa-NS  &            	
    \underline{0.5968}  &   32.44     &     
    0.5244        &         42.99        & 
    \underline{0.6236}        &         39.65                 \\ \midrule
\multirow{4}{*}{Ours}   & BERT-GCN          &                           	0.7075      &    69.26	 &
        \textbf{0.6312}        &             58.13            & 
        \textbf{0.6888}             &          63.75               \\
    & RoBERTa-GCN       &
    \textbf{0.7221} &          69.04               &
    0.6233      &             61.44        	   &
    0.6802      &   65.73              \\
    & BERT-GAT          &
    0.5585 &   61.93                      &
    0.4553 &   41.93	& 	
    0.4568                      & 62.18               \\
    & RoBERTa-GAT       &                            
    0.5692&               64.51          & 
    0.4358&               24.74  	        &                           0.4585  &            59.55             \\ \bottomrule 
\end{tabular}
\caption{Comparison with Baselines on Best Test Area under Precision-Recall Curve (PRAUC) and its corresponding F1 for  \textbf{CTFW} (CTFwrite-up), \textbf{COR} (Cooking), \textbf{MAM} (Maintenance) datasets.  NS is the next sentence based prediction approach. Our best model performance is bold, while maximum baseline performance is underlined.}
\label{tab:main_result}
\end{table*}

\noindent
\textbf{Metrics:}
We use \textit{accuracy} as the evaluation metric for the Sentence Type classification task on CTFW.
For the second task, because of the label imbalance we compare based on the \textit{area under Precision-Recall curve} (PRAUC) and also report the corresponding \textit{F1-score}. 
Hence do not report area under the ROC curve or accuracy. 

We consider four settings for this task.  The \textit{no window} setting ($W_{all}$) checks whether there is an edge between any two statements in the given document. The comparisons required in this setting are directly proportional to the document's size. In CTFW, the size of each write-up is quite large. So, to reduce unnecessary comparisons, we apply simple heuristics that instructions in procedural text, in general, does not have longer \textit{direct} dependencies. Thus, using the windows, we can control each sentence's number of comparisons (node). To understand how the performances change we evaluate with a \textit{sliding windows of N sentences} ($W_N$)  where $N = 3, 4, 5$. The comparisons are only made with the next $N$ sentences from a given sentence. For example, in case of $W_5$, for first sentence ($S_1$) we check for edges with $S_2$, $S_3$, $S_4$, $S_5$, $S_6$ and not $S_7$ on-wards. However, to have a fair comparison, we keep labeled out-of-window gold edges, if any. The ratios of existing and total edges in CTFW  are  0.07 ($W_{all}$), 0.24 ($W_5$), 0.29 ($W_4$), 0.38 ($W_3$).

\noindent
\textbf{Training:}
We use Pytorch Geometric \cite{fey2019fast} for GNN and transformers \cite{wolf2020transformers} for LM implementations.  Training is done with AdamW \cite{loshchilov2017decoupled} optimizer along with linear warmup scheduler on 4 Tesla V100 16GB GPUs.  We use bert-base-uncased, bert-large-uncased, roberta-base and roberta-large versions as  base model. We store the model with the best PRAUC score. Batch size of \{4,8,16\} and learning rates of \{1e-5,5e-6\} are used. Maximum sequence length varies between \{64, 80, 128\}. GNN depths are kept 128 (layer 1) and 64 (layer 2). We use a dropout of 0.4 in selected layers. For GAT, we keep four attention heads in layer 1. Details are present in Appendix \ref{sub:appendix:training}.

\section{Results and Discussion}


\subsection{Sentence Type Classification (STC)}
We use large and base versions of BERT and RoBERTa for this task to predict the type of sentences in a given text to establish a baseline for this task. This task  helps to identify relevant and irrelevant sentences in a document. Each sentence is classified into any of Action, Information, Both, Code, and None. These fine-grained annotations can be used in later  works for creating automated agents for vulnerability analysis.  The processed data consists of 120751 samples for training, 17331 for validation, and 34263 for testing. 
Table \ref{tab:sent_cls} shows that  RoBERTa-large  performs better than the rest.

\begin{table}[H]
\centering
\small
\begin{tabular}{@{}lcc@{}}
\toprule
\textbf{Model}      & \textbf{Val}   & \textbf{Test}  \\ \midrule
BERT-Base  & 78.48$\pm$0.25 &	77.42$\pm$0.10  \\ 
BERT-Large & 78.19$\pm$0.48 &	77.13$\pm$0.20 \\
RoBERTa-Base  & 78.85$\pm$0.25 &	77.37$\pm$0.11  \\
RoBERTa-Large &  79.02$\pm$0.16 &	\textbf{77.66$\pm$0.12}  \\  \bottomrule
\end{tabular}
\caption{Sentence Type Classification (Mean Accuracy from three seed values). Best performance in bold.}
\label{tab:sent_cls}
\end{table}

\begin{figure*}[!htb]
  \centering

  \includegraphics[scale=0.18]{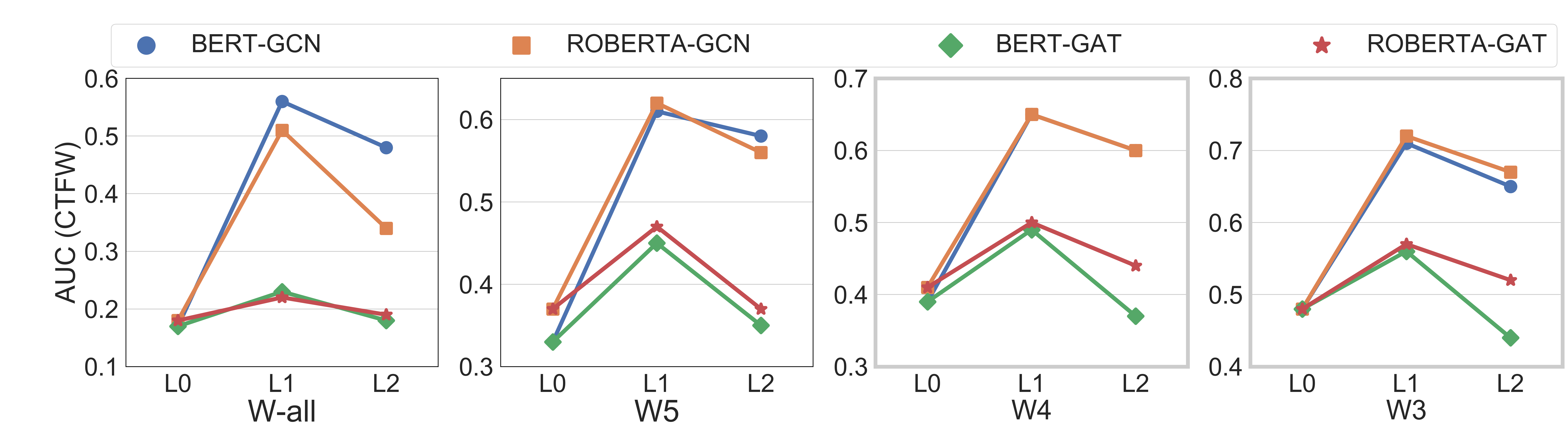}
  \includegraphics[scale=0.18,trim={5cm 0 0 0}]{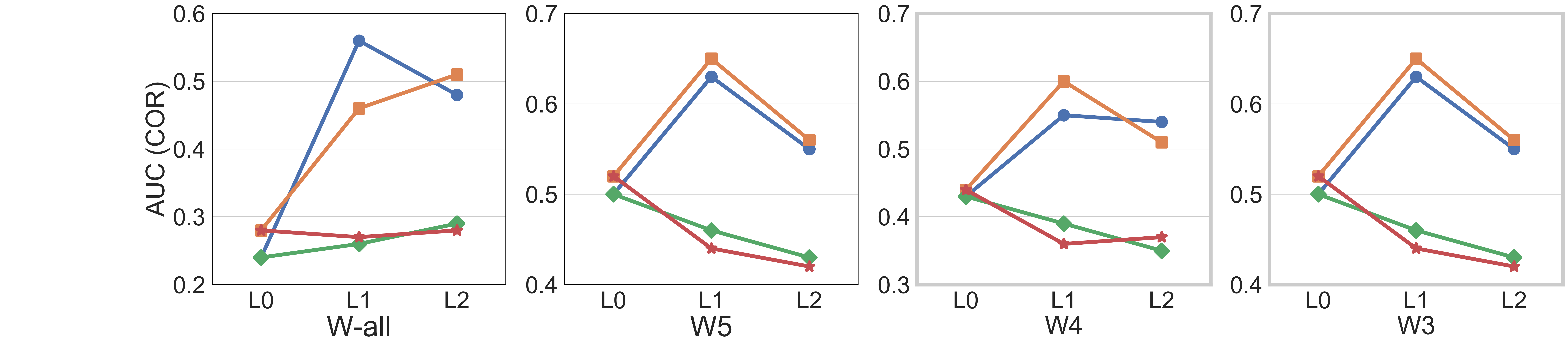}
  \includegraphics[scale=0.18,trim={5cm 0 0 0}]{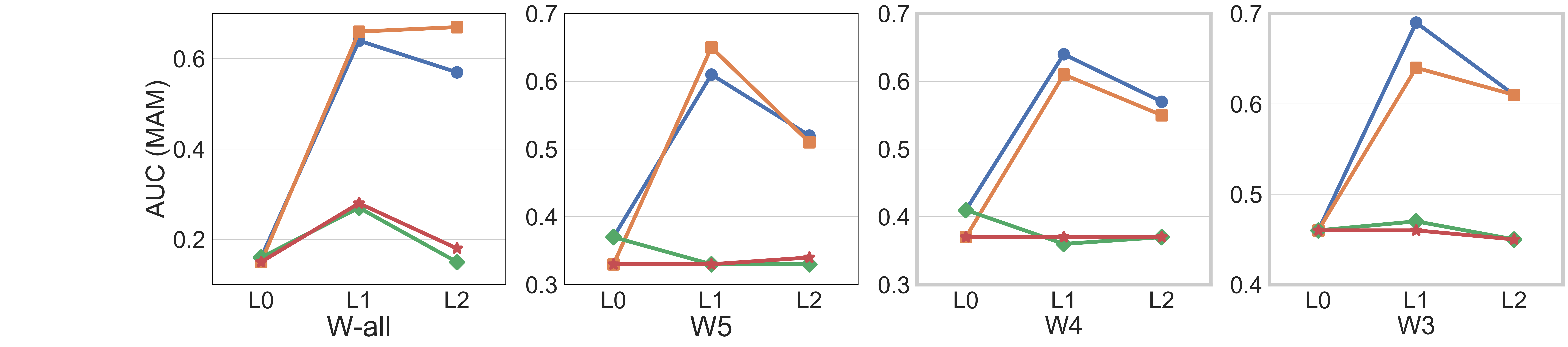}
  
  \caption{Effect of GNN Layers ($L_0$, $L_1$, $L_2$) on performance  (PRAUC) of the models for $W_{all}$, $W_5$, $W_4$, $W_3$ settings on the three datasets}
  \label{fig:CTFW_GNN_LAYERS}
\end{figure*}

\subsection{Flow Structure Prediction}
Here we present the performance results for the flow structure prediction. 

\noindent
\textbf{Random Baseline:}
In the \textit{Random} baseline, for every pair of nodes in each document we randomly select 0  (no-edge) or 1 (edge). For \textit{Weighted Random} baseline, we choose randomly, based on the percentage of edge present in the train set. We only report F1 since there is no probability calculation.

\noindent
\textbf{Next Sentence-based Prediction (NS) Baseline:}
We use LMs like BERT and RoBERTa in a next sentence prediction setting to get the baselines. Each pair of sentences is concatenated using [SEP] token and passed through these language models. Using the pooled LM representation, we classify whether an edge exists between them or not. We show maximum PRAUC and its corresponding F1 for each dataset from the results of each of our window settings ($W_3$, $W_4$, $W_5$, $W_{all}$).

\noindent
\textbf{Models:} 
We compare four variants of our LM-GNN models both with baseline and among each other in Table \ref{tab:main_result}. The scores are overall best scores across single and double layers GNN (GCN/GAT) and LM (BERT/RoBERTa) after experiments with both base and large version, trained with  pre-trained and randomly initialized weights.

We  see that the best LM-GCN models outperform the best baseline model by 0.12, 0.07, 0.06 in PRAUC for CTFW, COR, and MAM datasets, respectively. However, the best LM-GAT scores falls short of the baselines indicating that the graph attentions on LM \textit{sentence representations} cannot learn robust representation to perform this edge prediction task. Another thing to notice here is that, the best BERT-GCN models perform better than RoBERTa-GCN for COR and MAM datasets while performs poorly in the CTFW dataset. We hypothesize that this  is because, the CTFW dataset has ten times more data than COR and six times more than MAM, which helps the RoBERTa model correctly predict the edges.

\begin{table}[]
\centering
\small
\begin{tabular}{@{}lllll@{}}
\toprule
       & $W_3$ & $W_4$ & $W_5$ & $W_{all}$ \\
\midrule
CTFW-SC & 0.6630   &  0.5985  &  0.5733  &  \textbf{0.5590}    \\
CTFW-L & \textbf{0.7221}   &  \textbf{0.6520}  &  \textbf{0.6150}  &  0.3962    \\
CTFW-EP   &   0.3700  &   0.2900  &   0.2400  &   0.0700\\
\midrule
COR-SC  &  0.5639  &  0.5129  &  0.4731  &    \textbf{0.5580}  \\
COR-L  &  \textbf{0.6456}  &  \textbf{0.6012}  &   \textbf{0.5274} &    0.4034  \\
COR-EP   &   0.3700   &   0.3100   &   0.2600   &   0.1700\\
\midrule
MAM-SC  &  0.6528  &   0.6219 &   0.6091 &     \textbf{0.6718} \\
MAM-L  &   \textbf{0.6888} &  \textbf{0.6362}  &  \textbf{0.6137}  &  0.4161   \\
MAM-EP   &   0.4500   &   0.3700   &   0.3200   &   0.1500 \\
			
\bottomrule
\end{tabular}
\caption{Effect of  Semi-Complete(SC) and Linear(L) Graph Connection on 3 datasets in Area under Precision-Recall Curve (PRAUC). We also keep edge-percent (EP) in four window settings for comparison. }
\label{tab:effect_graphconn}
\end{table}

\subsection{Analysis}
\paragraph{Effect of Graph Connection Type:}
Table \ref{tab:effect_graphconn} shows how the models behave with semi-complete (SC) and linear (L) graph connection. For each dataset, we compare the PRAUC results for each window to draw more granular insight on the effect of neighbor aware representation learning. When we restrict graph learning by creating small windows ($W_3$, $W_4$, $W_5$), the linear model works better because of its selective learning conditioned on its predecessor. On the other hand, the semi-complete connection helps to learn a global awareness and works best in the $W_{all}$ setting. It is important to note that each model performs better than the average PRAUC performance, which is the percentage of edges in the data indicating that the model  is able to learn using the graph connections.

\noindent
\textbf{Effect of Graph Layers:}
We study how the depth of the GNNs affects the performance. We compare PRAUC across all four variations of the model in No-Window ($W_{all}$), $W_5$, $W_4$, $W_3$ settings in Figure \ref{fig:CTFW_GNN_LAYERS}. We experimented with no ($L_0$), single ($L_1$) and double ($L_2$) GNN layers. In all three datasets, we find the performance improves when we use a single layer and degrades beyond that for each of the windows with GCN based models. We do not go beyond two layers because of this observation and the graph connection types we use. We believe the reason for this drop (0.03-0.08 PRAUC) is that information from 2-hop neighbors might hinder the learning of the current node and confuse the model to predict wrongly. The GAT-based models mostly remain unaffected with the graph layers for both COR and MAM while showing some improvement in CTFW for one layer setting.

\begin{figure}[ht]
  \includegraphics[width=\linewidth]{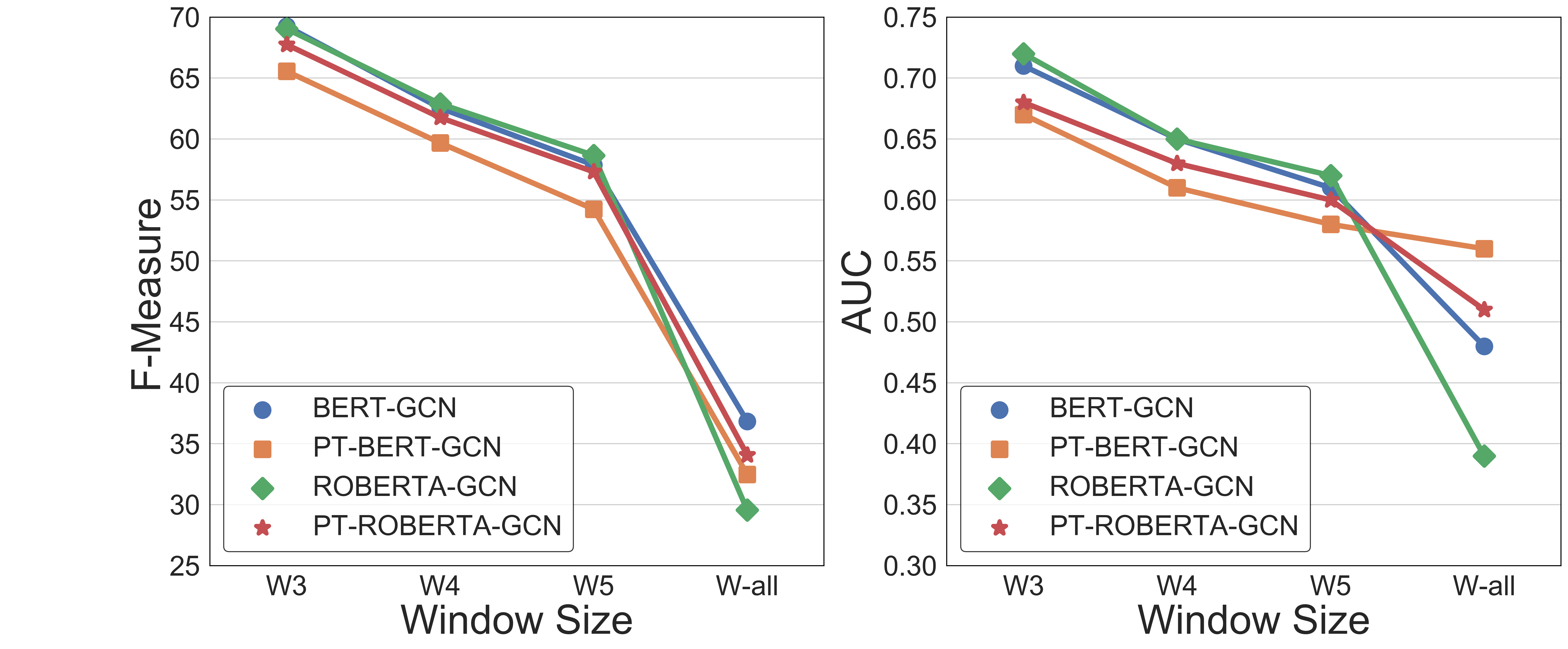}
  \includegraphics[width=\linewidth]{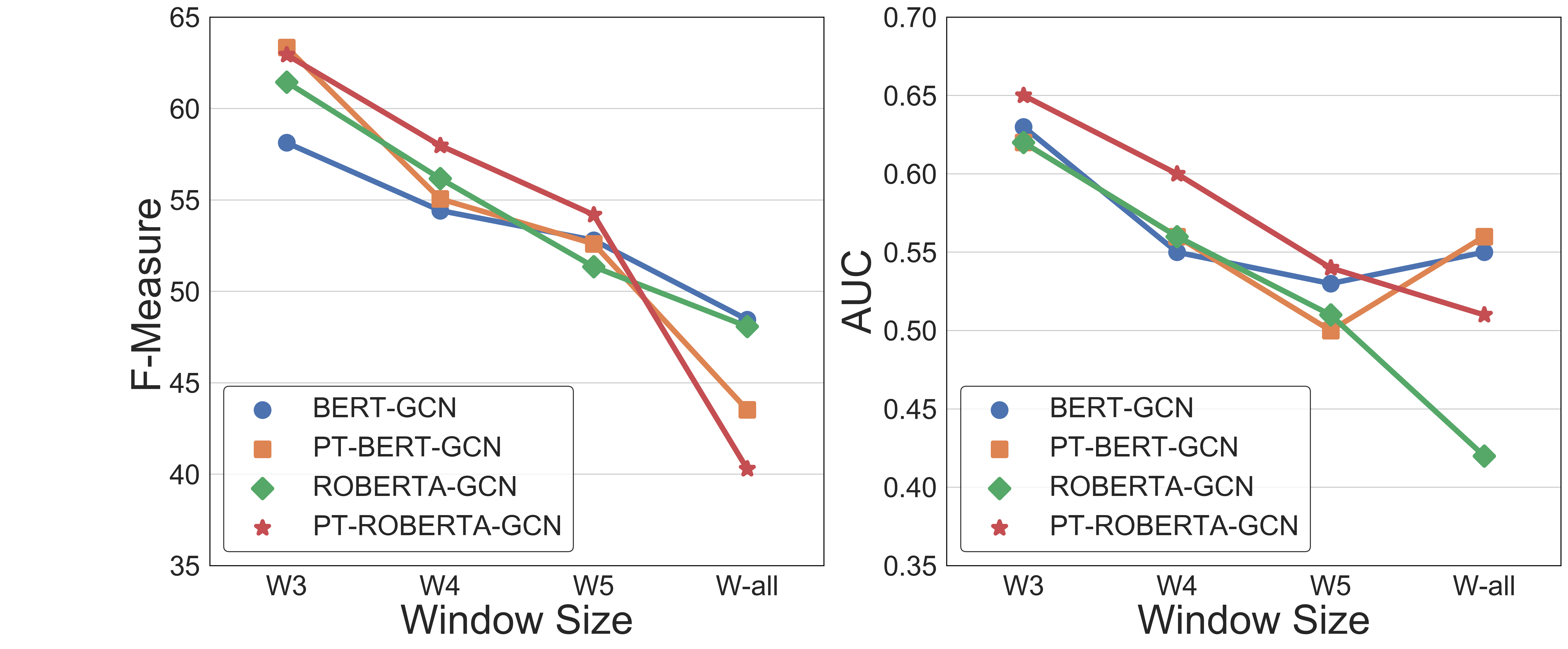}
  \includegraphics[width=\linewidth]{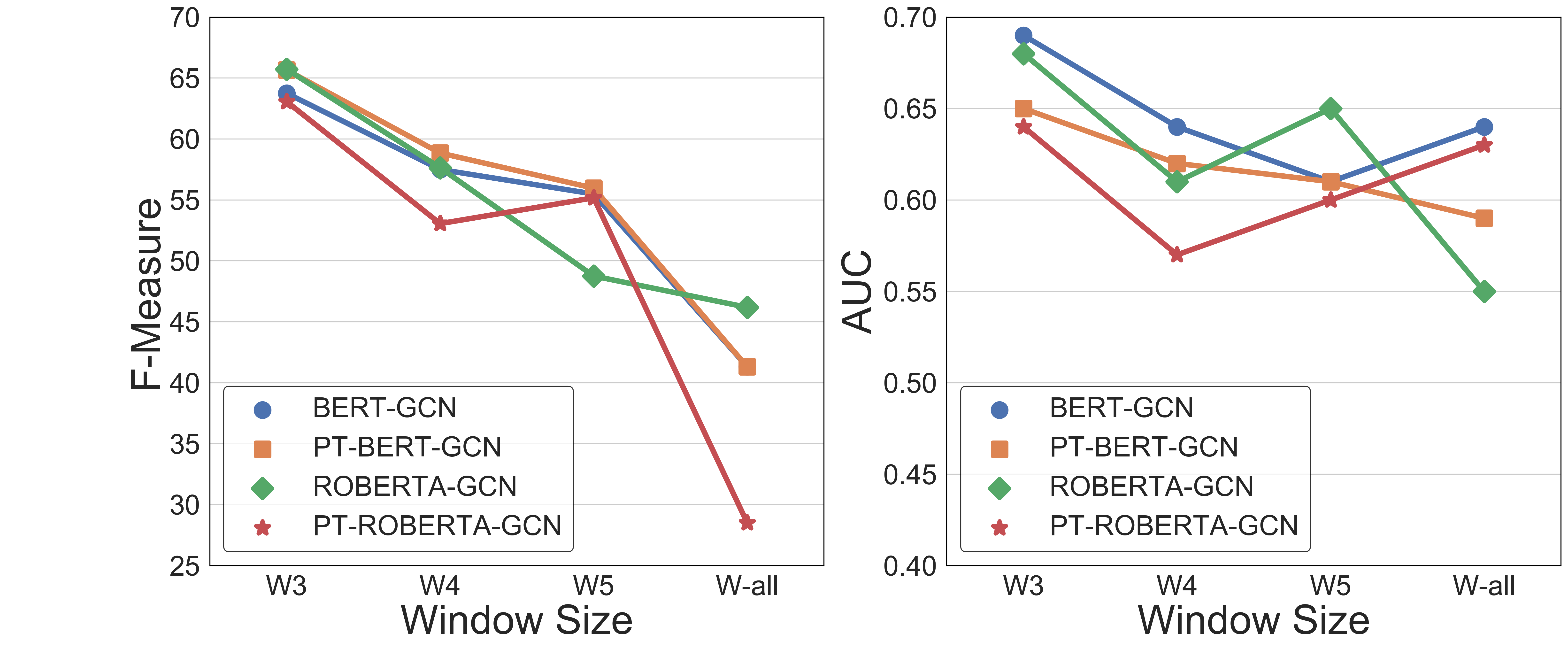}
  \caption{Performance for CTFW, COR, MAM trained from \textbf{scratch} and fine-tuned with \textbf{pre-trained weights}. }
  \label{fig:CTFW_PT}
\end{figure}




\noindent
\textbf{Effect of Pre-trained LM Weights:}
We study the impact of pre-trained weights of BERT and RoBERTa on the performance in Figure \ref{fig:CTFW_PT}. We notice, for the three datasets, the performance slightly decreases when the pre-trained model weights are used. This observation may be  because the texts' nature is quite different from the type of texts these LMs have been pre-trained on. The CTFW data often contains code fragments embedded in sentences, emoticons, or common conversational languages used in public forums.

\begin{figure}[]
  \includegraphics[width=\linewidth]{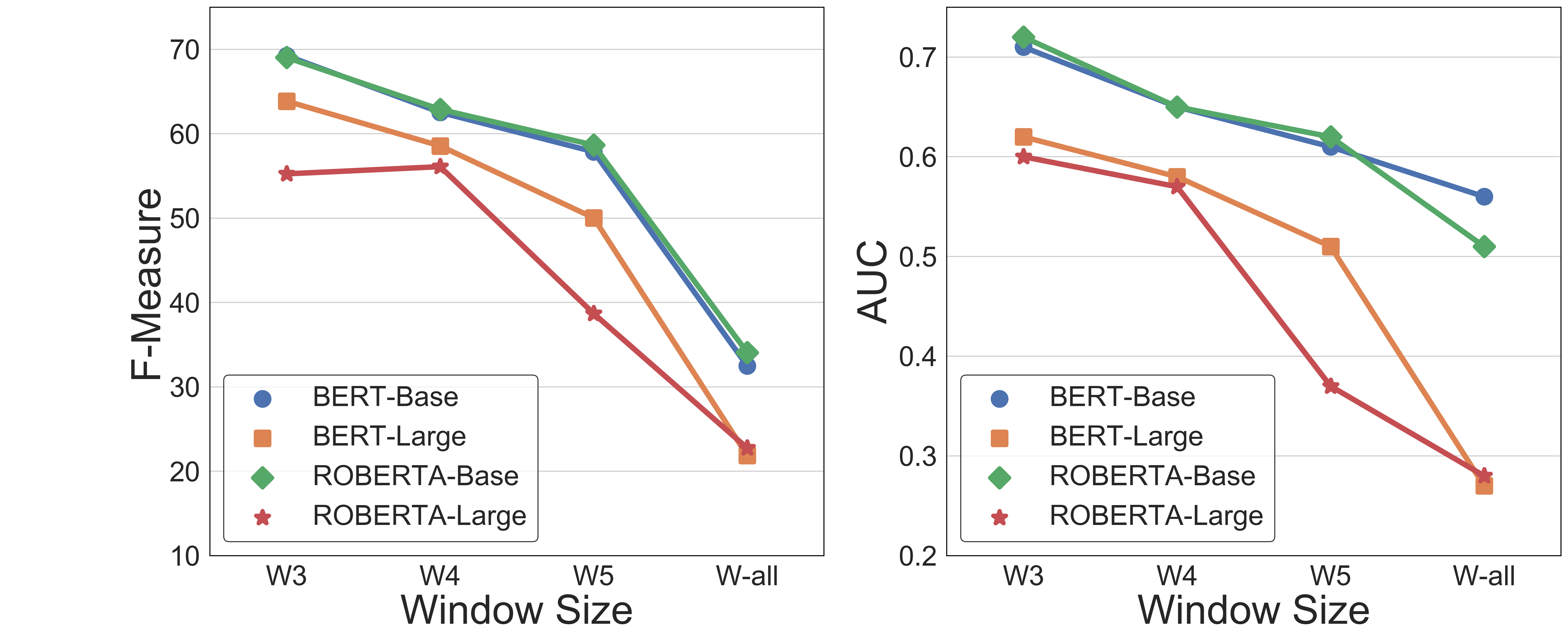}
  \includegraphics[width=\linewidth]{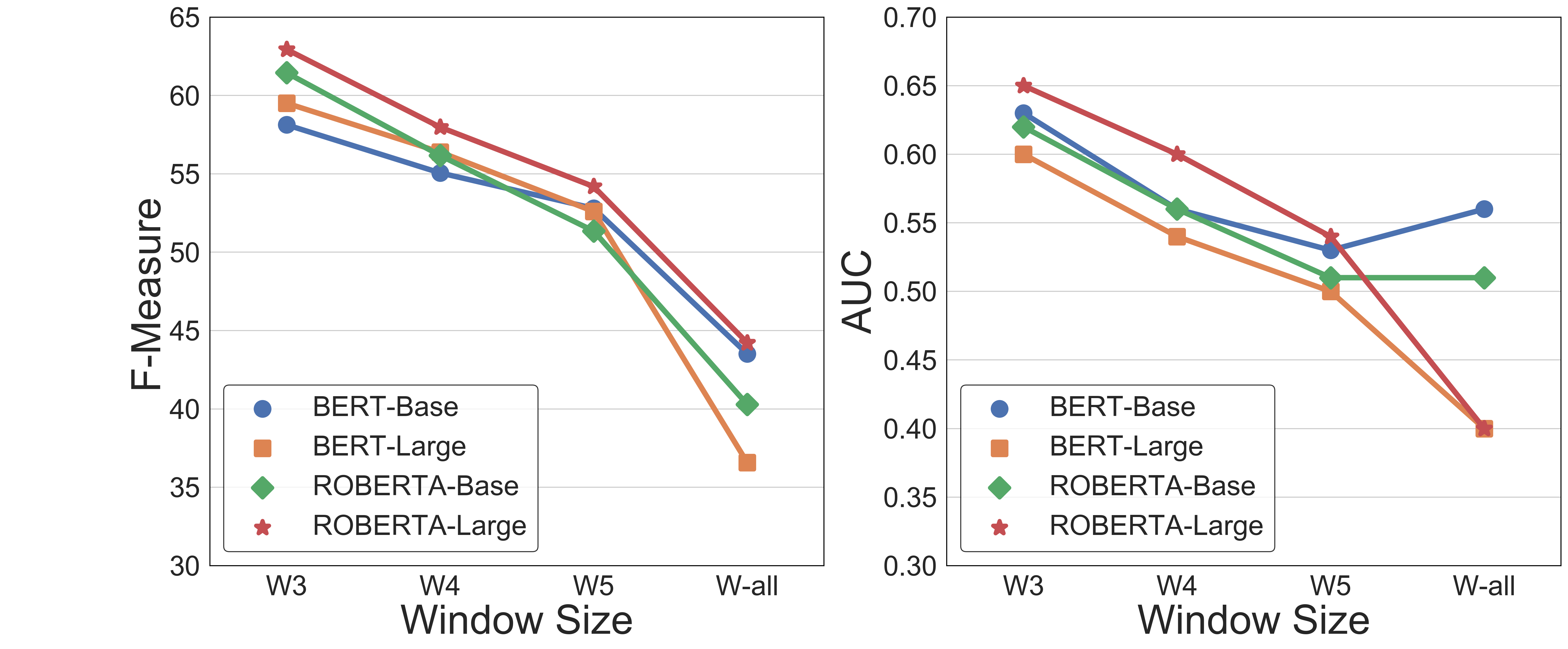}
  \includegraphics[width=\linewidth]{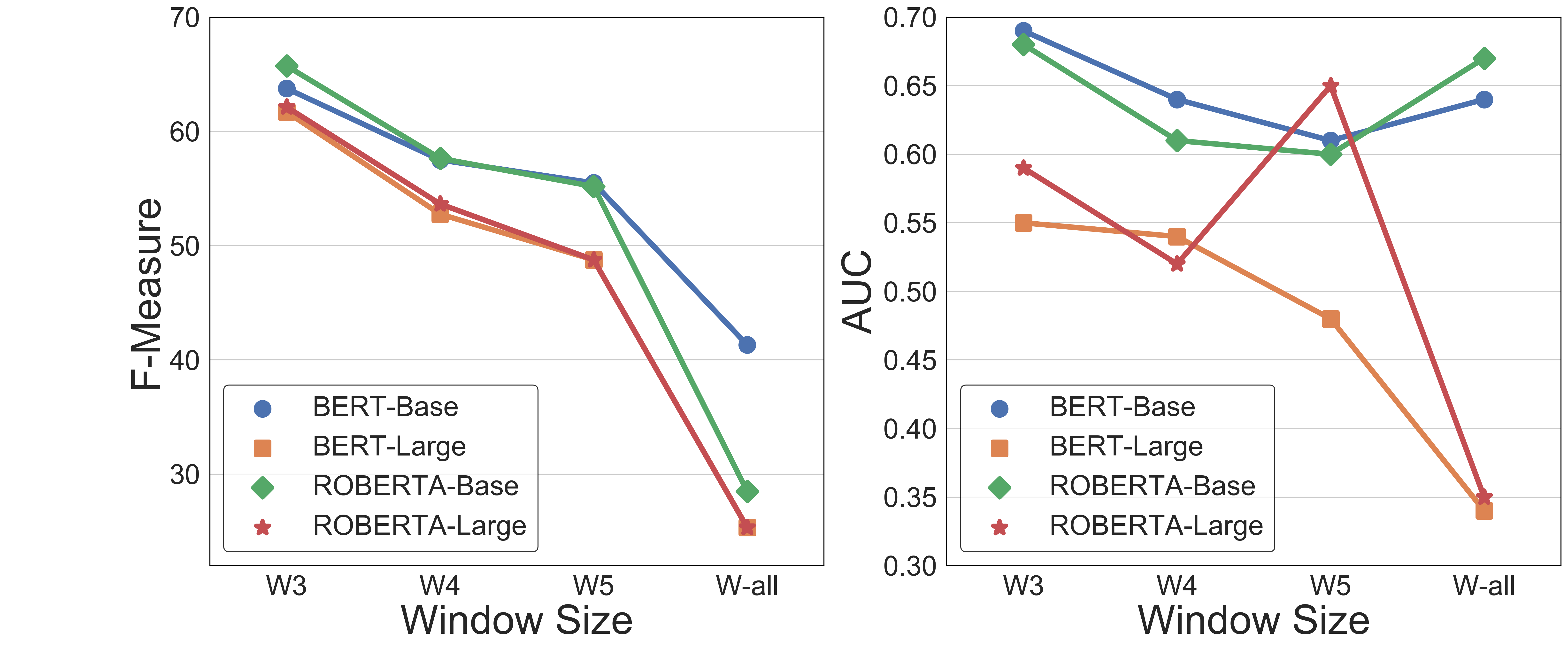}
  \caption{Performance on CTFW, COR, MAM trained with \textbf{base} and \textbf{large} version of the model. }
  \label{fig:CTFW_LARGE}
\end{figure}

\noindent
\textbf{Effect of LM Size:}
We also experimented with the size of sentence embeddings to see if that makes any difference to the performance. We use base and large version of BERT and RoBERTa for the experiments across  three datasets. We present the impact on F1 and PRAUC in Figure \ref{fig:CTFW_LARGE}. The performance of the larger versions of the  models drop in all three datasets. This drop, we believe, is because the sentences in these texts are relatively short and help the smaller versions of the models with lesser parameters to learn better.

\noindent
\textbf{Other experiments:}
We also experimented with modifications of other parts of the models like changing the number of projection layers, projection layer sizes, the number of attention heads in the GAT model, or dropout percent in selected layers and modes of message aggregation (add, max, mean). We do not report them since they do not significantly change PRAUC values.

\section{Related Work}

\noindent
\textbf{Procedural knowledge extraction:}
There are attempts to extract structured knowledge from cooking instructions in the form of named entities \cite{malmaud2014cooking}, their sentence-level dependencies \cite{mori2014flow, maeta2015framework, xu2020benchmark}, and action-verb argument flow across sentences \cite{jermsurawong2015predicting, kiddon2015mise,pan2020multi}.
In other domains, extraction of clinical steps from medline abstracts \cite{song2011procedural}, extraction of material synthesis operations and its arguments in material science \cite{mysore2019materials}, providing structures to how-to procedures \cite{park2018learning}, and action-argument retrieval from web design tutorials \cite{yang2019creative} mostly focus on fine-grained entity extractions rather than action or information traces.
The goal of our paper is constructing flow graphs from free-form, natural-language procedural texts without diverse domain knowledge.
Hence, we refrain from training specialized named-entity recognizers for each domain to find specific entities.
Our work is related to event or process discovery in process modeling tasks \cite{epure2015automatic, honkisz2018concept, qian2020approach, hanga2020graph}, but our goal is not finding specific events or actions from procedural texts.
In addition, the recent research proposed a method to create the forum structures from an unstructured forum based on the contents of each post using  BERT's Next Sentence Prediction \cite{kashihara2020social}. However, we focus on building flow graphs for procedural texts using GNNs.

\noindent
\textbf{Graph Neural Networks:}
GNNs are important in reasoning with graph-structured data in three major tasks, node classification \cite{kipf2016semi,hamilton2017inductive}, link prediction \cite{schlichtkrull2018modeling}, and graph classification \cite{ying2018hierarchical,pan2015joint,pan2016task,zhang2018end}.
GNNs help learn better node representations in each task using neural message passing \cite{gilmer2017neural} among connected neighbors.
We consider two widely used GNNs, GCN (Graph Convolutional Network) \cite{kipf2016semi} and GAT (Graph Attention Networks) \cite{velivckovic2017graph} to learn sentence representation to provide a better edge prediction.


\noindent
\textbf{Edge Prediction Task:}
Edge or link prediction tasks~\cite{li2018link, zhang2018link, pandey2019comprehensive, haonan2019graph,bacciu2019graph} work mainly on pre-existing networks or social graphs as inputs and predict the existence of future edges between nodes by extracting graph-specific features.
Different from existing work, we modeled the task of generating a graph-structure from a given natural-language text as an edge prediction task in a graph and learning representations of sentences considered as nodes. 

\noindent
\textbf{Combinations of BERT and GCN:}
 Recent works have used concatenation of BERT and GCN representations of texts or entities to improve performance of tasks like commonsense knowledge-base completion \cite{malaviya2019exploiting},  text classification \cite{yedocument, lu2020vgcn}, multi-hop reasoning  \cite{xiao2019dynamically}, citation recommendation \cite{jeong2019context}, medication recommendation \cite{shang2019pre}, relation extraction \cite{zhao2019improving}.
Graph-BERT \cite{zhang2020graph} solely depends on attention layers of BERT without using any message aggregation techniques. However, we differ from each of the previous methods in terms of model architecture, where we use BERT to learn initial sentence representations and GCN or GAT to improve them by learning representations from its neighboring connected sentences. BERT-GAT  for MRC \cite{zheng-etal-2020-document} created the graph structure from the well-structured wikipedia data whereas we explore two predefined natures of graph structures because of the free-formed text nature without such well-defined text-sections, presence of code-fragments, emoticons, and unrelated-token.

\section{Conclusion and Future Work}

We introduce a new procedural sentence flow extraction task from natural-language texts.
This task is important for procedural texts in every domain.
We create a sufficiently large procedural text dataset in the cybersecurity domain (CTFW) and construct structures from the natural form.
We empirically show that this task can be generalized across multiple domains with different natures and styles of texts.
In this paper, we only focus on English security write-ups.
As part of  future work, we plan to build automated agents in the cybersecurity domain to help and guide novices in performing software vulnerability analysis.
We also plan to include non-English write-ups.
We hope the CTFW dataset will facilitate other works in this research area.

\section*{Acknowledgement}
The authors acknowledge support from the Defense Advanced Research Projects Agency (DARPA) grant number FA875019C0003 for this project.

\clearpage
\section*{Impact Statement}
The dataset introduced here consists of write-ups written in public forums by students or security professionals from their personal experiences in the CTF challenges. The aggregated knowledge of such experiences is immense. This in-depth knowledge of the analysis tools and the approach to a problem is ideal for students working in software vulnerability analysis to learn from. Automated tutors built using such knowledge can reduce the efforts and time wasted in manually reading through a series of lengthy write-up documents.

CTFTime website states that the write-ups are copyrighted by the authors who posted them and it was practically impossible to contact each authors. It is also allowed to use the data for research purposes \cite{uscopyright, eucopyright}
Thus, we follow the previous work \cite{vsvabensky2021cybersecurity} using data from CTFTime and share only the urls of those write-ups from the CTFTime website which we use. We do not provide the scraper script since it would create a local copy of the write-up files unauthorized by the users. Interested readers can replicate the simple scraper script from the instructions in Appendix \ref{sub:appendix:dataset} and use it after reviewing the conditions under which it is permissible to use. We, however, share our annotations for those write-ups files.

Part of the annotations were provided as an optional, extra-credit assignment for the Information Assurance course. These CTF write-ups were directly related to the course-content, where students were required to read existing CTF write-ups and write write-ups for other security challenges they worked on during the course. Then students were given the option of voluntarily annotating CTF write-ups they read for extra credits in the course. For this task, we followed all the existing annotation guidelines and practices. We also ensured that  
\begin{itemize}
    \item The volunteers were aware of the fact that their annotations would be used for a research project.
    \item They were aware that no PII was involved or would be used in the research project.
    \item They were aware that extra credits were entirely optional, and they could refrain from submitting at any point of time without any consequences.
    \item Each volunteer was assigned only 10-15 write-ups based on a pilot study we did ahead of time, annotating an average-length CTF write-up took about two minutes (maximum ten mins).
\end{itemize}

Remaining annotations were performed by the Teaching Assistants (TA) of the course. These annotations were done as part of the course preparation process, which was part of their work contract. All the TAs were paid bi-weekly compensation by the university or by research funding. It was also ensured that the TAs knew these annotations would be used for a research project, their PII was not involved and annotations were to be anonymized before using.


\bibliographystyle{acl_natbib}
\bibliography{camera_ready}

\begin{thebibliography}{55}
\expandafter\ifx\csname natexlab\endcsname\relax\def\natexlab#1{#1}\fi

\bibitem[{euc()}]{eucopyright}

\newblock European union, copyright in the eu, 2020.
\newblock
  \url{https://europa.eu/youreurope/business/running-business/intellectual-property/copyright/index_en.htm}.
\newblock Accessed: 2021-05-01.

\bibitem[{usc()}]{uscopyright}

\newblock Us copyright office, copyright law of the united states, 2016.
\newblock \url{https://www.copyright.gov/title17/92chap1.html#107}.
\newblock Accessed: 2021-05-01.

\bibitem[{Bacciu et~al.(2019)Bacciu, Micheli, and Podda}]{bacciu2019graph}
Davide Bacciu, Alessio Micheli, and Marco Podda. 2019.
\newblock Graph generation by sequential edge prediction.
\newblock In \emph{ESANN}.

\bibitem[{{CTFTime}()}]{ctftime}
{CTFTime}.
\newblock {CTFTime}.
\newblock \url{https://ctftime.org}.
\newblock Accessed: 2021-05-01.

\bibitem[{Delpech and
  Saint-Dizier(2008)}]{delpech-saint-dizier-2008-investigating}
Estelle Delpech and Patrick Saint-Dizier. 2008.
\newblock \href
  {http://www.lrec-conf.org/proceedings/lrec2008/pdf/20_paper.pdf}
  {Investigating the structure of procedural texts for answering how-to
  questions}.
\newblock In \emph{Proceedings of the Sixth International Conference on
  Language Resources and Evaluation ({LREC}'08)}, Marrakech, Morocco. European
  Language Resources Association (ELRA).

\bibitem[{Devlin et~al.(2018)Devlin, Chang, Lee, and
  Toutanova}]{devlin2018bert}
Jacob Devlin, Ming-Wei Chang, Kenton Lee, and Kristina Toutanova. 2018.
\newblock Bert: Pre-training of deep bidirectional transformers for language
  understanding.
\newblock \emph{arXiv preprint arXiv:1810.04805}.

\bibitem[{Epure et~al.(2015)Epure, Mart{\'\i}n-Rodilla, Hug, Deneck{\`e}re, and
  Salinesi}]{epure2015automatic}
Elena~Viorica Epure, Patricia Mart{\'\i}n-Rodilla, Charlotte Hug, Rebecca
  Deneck{\`e}re, and Camille Salinesi. 2015.
\newblock Automatic process model discovery from textual methodologies.
\newblock In \emph{2015 IEEE 9th International Conference on Research
  Challenges in Information Science (RCIS)}, pages 19--30. IEEE.

\bibitem[{Fey and Lenssen(2019)}]{fey2019fast}
Matthias Fey and Jan~Eric Lenssen. 2019.
\newblock Fast graph representation learning with pytorch geometric.
\newblock \emph{arXiv preprint arXiv:1903.02428}.

\bibitem[{Fontan and Saint-Dizier(2008)}]{fontan2008analyzing}
Lionel Fontan and Patrick Saint-Dizier. 2008.
\newblock Analyzing the explanation structure of procedural texts: Dealing with
  advice and warnings.
\newblock In \emph{Semantics in Text Processing. STEP 2008 Conference
  Proceedings}, pages 115--127.

\bibitem[{Gilmer et~al.(2017)Gilmer, Schoenholz, Riley, Vinyals, and
  Dahl}]{gilmer2017neural}
Justin Gilmer, Samuel~S Schoenholz, Patrick~F Riley, Oriol Vinyals, and
  George~E Dahl. 2017.
\newblock Neural message passing for quantum chemistry.
\newblock \emph{arXiv preprint arXiv:1704.01212}.

\bibitem[{Hamilton et~al.(2017)Hamilton, Ying, and
  Leskovec}]{hamilton2017inductive}
Will Hamilton, Zhitao Ying, and Jure Leskovec. 2017.
\newblock Inductive representation learning on large graphs.
\newblock In \emph{Advances in neural information processing systems}, pages
  1024--1034.

\bibitem[{Hanga et~al.(2020)Hanga, Kovalchuk, and Gaber}]{hanga2020graph}
Khadijah~Muzzammil Hanga, Yevgeniya Kovalchuk, and Mohamed~Medhat Gaber. 2020.
\newblock A graph-based approach to interpreting recurrent neural networks in
  process mining.
\newblock \emph{IEEE Access}, 8:172923--172938.

\bibitem[{Haonan et~al.(2019)Haonan, Huang, Ye, and Xiuyan}]{haonan2019graph}
Lu~Haonan, Seth~H Huang, Tian Ye, and Guo Xiuyan. 2019.
\newblock Graph star net for generalized multi-task learning.
\newblock \emph{arXiv preprint arXiv:1906.12330}.

\bibitem[{Hendrycks and Gimpel(2016)}]{hendrycks2016gaussian}
Dan Hendrycks and Kevin Gimpel. 2016.
\newblock Gaussian error linear units (gelus).
\newblock \emph{arXiv preprint arXiv:1606.08415}.

\bibitem[{Honkisz et~al.(2018)Honkisz, Kluza, and
  Wi{\'s}niewski}]{honkisz2018concept}
Krzysztof Honkisz, Krzysztof Kluza, and Piotr Wi{\'s}niewski. 2018.
\newblock A concept for generating business process models from natural
  language description.
\newblock In \emph{International Conference on Knowledge Science, Engineering
  and Management}, pages 91--103. Springer.

\bibitem[{Jeong et~al.(2019)Jeong, Jang, Shin, Park, and
  Choi}]{jeong2019context}
Chanwoo Jeong, Sion Jang, Hyuna Shin, Eunjeong Park, and Sungchul Choi. 2019.
\newblock A context-aware citation recommendation model with bert and graph
  convolutional networks.
\newblock \emph{arXiv preprint arXiv:1903.06464}.

\bibitem[{Jermsurawong and Habash(2015)}]{jermsurawong2015predicting}
Jermsak Jermsurawong and Nizar Habash. 2015.
\newblock Predicting the structure of cooking recipes.
\newblock In \emph{Proceedings of the 2015 Conference on Empirical Methods in
  Natural Language Processing}, pages 781--786.

\bibitem[{Kashihara et~al.(2020)Kashihara, Shakarian, and
  Baral}]{kashihara2020social}
Kazuaki Kashihara, Jana Shakarian, and Chitta Baral. 2020.
\newblock Social structure construction from the forums using interaction
  coherence.
\newblock In \emph{Proceedings of the Future Technologies Conference}, pages
  830--843.

\bibitem[{Kiddon et~al.(2015)Kiddon, Ponnuraj, Zettlemoyer, and
  Choi}]{kiddon2015mise}
Chlo{\'e} Kiddon, Ganesa~Thandavam Ponnuraj, Luke Zettlemoyer, and Yejin Choi.
  2015.
\newblock Mise en place: Unsupervised interpretation of instructional recipes.
\newblock In \emph{Proceedings of the 2015 Conference on Empirical Methods in
  Natural Language Processing}, pages 982--992.

\bibitem[{Kipf and Welling(2016)}]{kipf2016semi}
Thomas~N Kipf and Max Welling. 2016.
\newblock Semi-supervised classification with graph convolutional networks.
\newblock \emph{arXiv preprint arXiv:1609.02907}.

\bibitem[{Li et~al.(2018)Li, Zhao, Ge, Yang, and Chen}]{li2018link}
Ji-chao Li, Dan-ling Zhao, Bing-Feng Ge, Ke-Wei Yang, and Ying-Wu Chen. 2018.
\newblock A link prediction method for heterogeneous networks based on bp
  neural network.
\newblock \emph{Physica A: Statistical Mechanics and its Applications},
  495:1--17.

\bibitem[{Liu et~al.(2019)Liu, Ott, Goyal, Du, Joshi, Chen, Levy, Lewis,
  Zettlemoyer, and Stoyanov}]{liu2019roberta}
Yinhan Liu, Myle Ott, Naman Goyal, Jingfei Du, Mandar Joshi, Danqi Chen, Omer
  Levy, Mike Lewis, Luke Zettlemoyer, and Veselin Stoyanov. 2019.
\newblock Roberta: A robustly optimized bert pretraining approach.
\newblock \emph{arXiv preprint arXiv:1907.11692}.

\bibitem[{Loshchilov and Hutter(2017)}]{loshchilov2017decoupled}
Ilya Loshchilov and Frank Hutter. 2017.
\newblock Decoupled weight decay regularization.
\newblock \emph{arXiv preprint arXiv:1711.05101}.

\bibitem[{Lu et~al.(2020)Lu, Du, and Nie}]{lu2020vgcn}
Zhibin Lu, Pan Du, and Jian-Yun Nie. 2020.
\newblock Vgcn-bert: Augmenting bert with graph embedding for text
  classification.
\newblock In \emph{European Conference on Information Retrieval}, pages
  369--382. Springer.

\bibitem[{Maeta et~al.(2015)Maeta, Sasada, and Mori}]{maeta2015framework}
Hirokuni Maeta, Tetsuro Sasada, and Shinsuke Mori. 2015.
\newblock A framework for procedural text understanding.
\newblock In \emph{Proceedings of the 14th International Conference on Parsing
  Technologies}, pages 50--60.

\bibitem[{Malaviya et~al.(2019)Malaviya, Bhagavatula, Bosselut, and
  Choi}]{malaviya2019exploiting}
Chaitanya Malaviya, Chandra Bhagavatula, Antoine Bosselut, and Yejin Choi.
  2019.
\newblock Exploiting structural and semantic context for commonsense knowledge
  base completion.
\newblock \emph{arXiv preprint arXiv:1910.02915}.

\bibitem[{Malmaud et~al.(2014)Malmaud, Wagner, Chang, and
  Murphy}]{malmaud2014cooking}
Jonathan Malmaud, Earl Wagner, Nancy Chang, and Kevin Murphy. 2014.
\newblock Cooking with semantics.
\newblock In \emph{Proceedings of the ACL 2014 Workshop on Semantic Parsing},
  pages 33--38.

\bibitem[{Mori et~al.(2014)Mori, Maeta, Yamakata, and Sasada}]{mori2014flow}
Shinsuke Mori, Hirokuni Maeta, Yoko Yamakata, and Tetsuro Sasada. 2014.
\newblock Flow graph corpus from recipe texts.
\newblock In \emph{LREC}, pages 2370--2377.

\bibitem[{Mysore et~al.(2019)Mysore, Jensen, Kim, Huang, Chang, Strubell,
  Flanigan, McCallum, and Olivetti}]{mysore2019materials}
Sheshera Mysore, Zach Jensen, Edward Kim, Kevin Huang, Haw-Shiuan Chang, Emma
  Strubell, Jeffrey Flanigan, Andrew McCallum, and Elsa Olivetti. 2019.
\newblock The materials science procedural text corpus: Annotating materials
  synthesis procedures with shallow semantic structures.
\newblock \emph{arXiv preprint arXiv:1905.06939}.

\bibitem[{Pan et~al.(2020)Pan, Chen, Wu, Liu, Ngo, Kan, Jiang, and
  Chua}]{pan2020multi}
Liang-Ming Pan, Jingjing Chen, Jianlong Wu, Shaoteng Liu, Chong-Wah Ngo,
  Min-Yen Kan, Yugang Jiang, and Tat-Seng Chua. 2020.
\newblock Multi-modal cooking workflow construction for food recipes.
\newblock In \emph{Proceedings of the 28th ACM International Conference on
  Multimedia}, pages 1132--1141.

\bibitem[{Pan et~al.(2016)Pan, Wu, Zhu, Long, and Zhang}]{pan2016task}
Shirui Pan, Jia Wu, Xingquan Zhu, Guodong Long, and Chengqi Zhang. 2016.
\newblock Task sensitive feature exploration and learning for multitask graph
  classification.
\newblock \emph{IEEE transactions on cybernetics}, 47(3):744--758.

\bibitem[{Pan et~al.(2015)Pan, Wu, Zhu, Zhang, and Philip}]{pan2015joint}
Shirui Pan, Jia Wu, Xingquan Zhu, Chengqi Zhang, and S~Yu Philip. 2015.
\newblock Joint structure feature exploration and regularization for multi-task
  graph classification.
\newblock \emph{IEEE Transactions on Knowledge and Data Engineering},
  28(3):715--728.

\bibitem[{Pandey et~al.(2019)Pandey, Bhanodia, Khamparia, and
  Pandey}]{pandey2019comprehensive}
Babita Pandey, Praveen~Kumar Bhanodia, Aditya Khamparia, and Devendra~Kumar
  Pandey. 2019.
\newblock A comprehensive survey of edge prediction in social networks:
  Techniques, parameters and challenges.
\newblock \emph{Expert Systems with Applications}, 124:164--181.

\bibitem[{Park and Motahari~Nezhad(2018)}]{park2018learning}
Hogun Park and Hamid~Reza Motahari~Nezhad. 2018.
\newblock Learning procedures from text: Codifying how-to procedures in deep
  neural networks.
\newblock In \emph{Companion Proceedings of the The Web Conference 2018}, pages
  351--358.

\bibitem[{Qian et~al.(2020)Qian, Wen, Kumar, Lin, Lin, Zong, Wang
  et~al.}]{qian2020approach}
Chen Qian, Lijie Wen, Akhil Kumar, Leilei Lin, Li~Lin, Zan Zong, Jianmin Wang,
  et~al. 2020.
\newblock An approach for process model extraction by multi-grained text
  classification.
\newblock In \emph{International Conference on Advanced Information Systems
  Engineering}, pages 268--282. Springer.

\bibitem[{Reitz()}]{pythonRequests}
Kenneth Reitz.
\newblock Requests: Http for humans.
\newblock \url{https://requests.readthedocs.io/en/master/}.
\newblock Accessed: 2020-10-23.

\bibitem[{Richardson(2007)}]{richardson2007beautiful}
Leonard Richardson. 2007.
\newblock Beautiful soup documentation.
\newblock \emph{April}.

\bibitem[{Schlichtkrull et~al.(2018)Schlichtkrull, Kipf, Bloem, Van Den~Berg,
  Titov, and Welling}]{schlichtkrull2018modeling}
Michael Schlichtkrull, Thomas~N Kipf, Peter Bloem, Rianne Van Den~Berg, Ivan
  Titov, and Max Welling. 2018.
\newblock Modeling relational data with graph convolutional networks.
\newblock In \emph{European Semantic Web Conference}, pages 593--607. Springer.

\bibitem[{Shang et~al.(2019)Shang, Ma, Xiao, and Sun}]{shang2019pre}
Junyuan Shang, Tengfei Ma, Cao Xiao, and Jimeng Sun. 2019.
\newblock Pre-training of graph augmented transformers for medication
  recommendation.
\newblock \emph{arXiv preprint arXiv:1906.00346}.

\bibitem[{Song et~al.(2011)Song, Oh, Myaeng, Choi, Chun, Choi, and
  Jeong}]{song2011procedural}
Sa-kwang Song, Heung-seon Oh, Sung~Hyon Myaeng, Sung-Pil Choi, Hong-Woo Chun,
  Yun-Soo Choi, and Chang-Hoo Jeong. 2011.
\newblock Procedural knowledge extraction on medline abstracts.
\newblock In \emph{International Conference on Active Media Technology}, pages
  345--354. Springer.

\bibitem[{spaCy(2017)}]{spaCy}
spaCy. 2017.
\newblock spacy v2.0.
\newblock \url{https://spacy.io/models/en#en_core_web_md}.

\bibitem[{{\v{S}}v{\'a}bensk{\`y} et~al.(2021){\v{S}}v{\'a}bensk{\`y},
  {\v{C}}eleda, Vykopal, and
  Bri{\v{s}}{\'a}kov{\'a}}]{vsvabensky2021cybersecurity}
Valdemar {\v{S}}v{\'a}bensk{\`y}, Pavel {\v{C}}eleda, Jan Vykopal, and Silvia
  Bri{\v{s}}{\'a}kov{\'a}. 2021.
\newblock Cybersecurity knowledge and skills taught in capture the flag
  challenges.
\newblock \emph{Computers \& Security}, 102:102154.

\bibitem[{Veli{\v{c}}kovi{\'c} et~al.(2017)Veli{\v{c}}kovi{\'c}, Cucurull,
  Casanova, Romero, Lio, and Bengio}]{velivckovic2017graph}
Petar Veli{\v{c}}kovi{\'c}, Guillem Cucurull, Arantxa Casanova, Adriana Romero,
  Pietro Lio, and Yoshua Bengio. 2017.
\newblock Graph attention networks.
\newblock \emph{arXiv preprint arXiv:1710.10903}.

\bibitem[{Wolf et~al.(2020)Wolf, Chaumond, Debut, Sanh, Delangue, Moi, Cistac,
  Funtowicz, Davison, Shleifer et~al.}]{wolf2020transformers}
Thomas Wolf, Julien Chaumond, Lysandre Debut, Victor Sanh, Clement Delangue,
  Anthony Moi, Pierric Cistac, Morgan Funtowicz, Joe Davison, Sam Shleifer,
  et~al. 2020.
\newblock Transformers: State-of-the-art natural language processing.
\newblock In \emph{Proceedings of the 2020 Conference on Empirical Methods in
  Natural Language Processing: System Demonstrations}, pages 38--45.

\bibitem[{Xiao et~al.(2019)Xiao, Qu, Qiu, Zhou, Li, Zhang, and
  Yu}]{xiao2019dynamically}
Yunxuan Xiao, Yanru Qu, Lin Qiu, Hao Zhou, Lei Li, Weinan Zhang, and Yong Yu.
  2019.
\newblock Dynamically fused graph network for multi-hop reasoning.
\newblock \emph{arXiv preprint arXiv:1905.06933}.

\bibitem[{Xu et~al.(2020)Xu, Ji, Shi, Du, Neubig, Bisk, and
  Duan}]{xu2020benchmark}
Frank~F Xu, Lei Ji, Botian Shi, Junyi Du, Graham Neubig, Yonatan Bisk, and Nan
  Duan. 2020.
\newblock A benchmark for structured procedural knowledge extraction from
  cooking videos.
\newblock \emph{arXiv preprint arXiv:2005.00706}.

\bibitem[{Yamakata et~al.(2020)Yamakata, Mori, and
  Carroll}]{yamakata2020english}
Yoko Yamakata, Shinsuke Mori, and John~A Carroll. 2020.
\newblock English recipe flow graph corpus.
\newblock In \emph{Proceedings of The 12th Language Resources and Evaluation
  Conference}, pages 5187--5194.

\bibitem[{Yang et~al.(2019)Yang, Fang, Jin, Chang, and
  Estrin}]{yang2019creative}
Longqi Yang, Chen Fang, Hailin Jin, Walter Chang, and Deborah Estrin. 2019.
\newblock Creative procedural-knowledge extraction from web design tutorials.
\newblock \emph{arXiv preprint arXiv:1904.08587}.

\bibitem[{Ye et~al.()Ye, Jiang, Liu, Li, and Yuan}]{yedocument}
Zhihao Ye, Gongyao Jiang, Ye~Liu, Zhiyong Li, and Jin Yuan.
\newblock Document and word representations generated by graph convolutional
  network and bert for short text classification.

\bibitem[{Ying et~al.(2018)Ying, You, Morris, Ren, Hamilton, and
  Leskovec}]{ying2018hierarchical}
Zhitao Ying, Jiaxuan You, Christopher Morris, Xiang Ren, Will Hamilton, and
  Jure Leskovec. 2018.
\newblock Hierarchical graph representation learning with differentiable
  pooling.
\newblock In \emph{Advances in neural information processing systems}, pages
  4800--4810.

\bibitem[{Zhang et~al.(2020)Zhang, Zhang, Sun, and Xia}]{zhang2020graph}
Jiawei Zhang, Haopeng Zhang, Li~Sun, and Congying Xia. 2020.
\newblock Graph-bert: Only attention is needed for learning graph
  representations.
\newblock \emph{arXiv preprint arXiv:2001.05140}.

\bibitem[{Zhang and Chen(2018)}]{zhang2018link}
Muhan Zhang and Yixin Chen. 2018.
\newblock Link prediction based on graph neural networks.
\newblock In \emph{Advances in Neural Information Processing Systems}, pages
  5165--5175.

\bibitem[{Zhang et~al.(2018)Zhang, Cui, Neumann, and Chen}]{zhang2018end}
Muhan Zhang, Zhicheng Cui, Marion Neumann, and Yixin Chen. 2018.
\newblock An end-to-end deep learning architecture for graph classification.
\newblock In \emph{Thirty-Second AAAI Conference on Artificial Intelligence}.

\bibitem[{Zhao et~al.(2019)Zhao, Wan, Gao, and Lin}]{zhao2019improving}
Yi~Zhao, Huaiyu Wan, Jianwei Gao, and Youfang Lin. 2019.
\newblock Improving relation classification by entity pair graph.
\newblock In \emph{Asian Conference on Machine Learning}, pages 1156--1171.

\bibitem[{Zheng et~al.(2020)Zheng, Wen, Liang, Duan, Che, Jiang, Zhou, and
  Liu}]{zheng-etal-2020-document}
Bo~Zheng, Haoyang Wen, Yaobo Liang, Nan Duan, Wanxiang Che, Daxin Jiang, Ming
  Zhou, and Ting Liu. 2020.
\newblock \href {https://doi.org/10.18653/v1/2020.acl-main.599} {Document
  modeling with graph attention networks for multi-grained machine reading
  comprehension}.
\newblock In \emph{Proceedings of the 58th Annual Meeting of the Association
  for Computational Linguistics}, pages 6708--6718, Online. Association for
  Computational Linguistics.

\end{thebibliography}
\clearpage
\appendix


\section{Extraction and Processing of Write-ups:}
\label{sub:appendix:dataset}
The extraction of CTF Write-up involved the following three phases.

\noindent
\textbf{Writeup URL extraction : }
We loop through all the write-up pages on ctftime website from page numbers 1 to 25500). We use a simple python scraper to scrape the content of each page using python  requests \cite{pythonRequests} library.
We look for keyword ``\textit{Original write-ups}" and extracted the href component if present. These URLs are stored for each writeup indexed with the page numbers.

\noindent
\textbf{Write-up Content extraction : }
We use these URLs and extract the contents of the write-ups using python libraries requests and BeautifulSoup \cite{richardson2007beautiful}. We extract all the text lines ignoring contents in html tags like style, scripts, head, title. The contents are stored in a text file named with the same page ids of the URLs.

\noindent
\textbf{Processing of Write-up : }
We processed and filter out sentences which do not have any verb forms using spacy \cite{spaCy} POS-Tagger. We cleaned and removed unnecessary spaces and split them into sentences. The processing script is available in the github.


\section{CTFW Data Statistics}
In CTFW, 
there are write-ups for 2236 unique tasks.  Only four out of those having more than 5 write-ups each. 72\% of the tasks have single write-up. The write-ups are from 311 unique competitions, ranging from years 2012-2019. A task having multiple write-ups vary in contents. In CTFW, only 3\% of the tasks have more than three write-ups, and 9\% have more than two. 

\section{Training Details:}
\label{sub:appendix:training}
 The correct set of hyperparameters are found by running three trials.  We run for \{50, 100\} epochs and store the model with the best PRAUC score. 
Each training with evaluation takes around 1-3 hours for base version of models and around 6 hours for larger versions depending upon the dataset used. The model parameters are directly proportional to the model parameters of language models, since the GNN only allow few more parameters as compared to the LMs.

\section{Baseline NS with weighted cross-entropy}
Table \ref{tab:wtdce_base} shows the PRAUC values when we use weighted cross-entropy with base version of BERT on unbalanced data during training. The results are not much different than the Next-Sentence baseline shown previously.
\begin{table}[!htb]
\begin{tabular}{@{}lllll@{}}
\toprule
\textbf{Dataset}     & \textbf{$W_3$} & \textbf{$W_4$} & \textbf{$W_5$} & \textbf{$W_{all}$} \\ \midrule
CTFW &  0.4613  & 0.4397   &  0.2546  &   0.3681   \\ 
COR  &  0.4724  & 0.4748   & 0.4837   &    0.4761  \\ 
MAM  &  0.5318  & 0.2318   & 0.2297   &    0.4724  \\ \bottomrule
\end{tabular}
\caption{BERT-base-uncased performance with NS prediction when Weighted Cross-Entropy used with Unbalanced Training Data}
\label{tab:wtdce_base}
\end{table}

\section{Number of comparisons Reduction using Windows}
We can control the total number of comparisons required to predict the edges in a graph by using the windows ($W_N$ where $N = 3, 4, 5, all$). The number of comparisons for each window is given by the equation \ref{eq:comp}. We can reduce the number of comparisons considerably for large documents using shorter windows of 3, 4, 5 sentences. The number of comparison $\mathcal{C}$ is defined by
\begin{align}
    \mathcal{C}=\Bigg\{\begin{matrix}
    max\{(n-s), 0\}s+\frac{s(s-1)}{2} & n=3,4,5\\
    {n \choose 2} & n=all
    \end{matrix}
    \label{eq:comp}
\end{align}

\section{CTFW STC Label statistics}
Table \ref{tab:sent_type} shows the label distributions of Sentence Type Classification data.
\begin{table}[ht]
\centering
\begin{tabular}{@{}lrrr@{}}

\toprule
\textbf{Label} & \textbf{Train} & \textbf{Val}  & \textbf{Test} \\ \midrule
A     & 11143  & 1499  & 3321  \\ 
I     & 23279  & 3075  & 6882 \\ 
A/I   & 2931   & 380   & 826  \\ 
C     & 1386  & 185  & 338  \\ 
NONE  & 82012  & 12192 & 22896 \\ \bottomrule
\end{tabular}
\caption{CTFW Sentence Type Classification}
\label{tab:sent_type}
\end{table}

\end{document}